**Reducing the labeling burden in time-series mapping using Common Ground: a semi-automated approach to tracking changes in land cover and species over time.**

Geethen Singh [a, b]*, Jasper A Slingsby [c, b], Tamara B Robinson [a], Glenn Moncrieff [b, d]

[a] Centre for Invasion Biology, Stellenbosch University, South Africa

[b] Fynbos Node, South African Environmental Observation Network, Observatory, Cape Town, South Africa

[c] Department of Biological Sciences and Centre for Statistics in Ecology, Environment and Conservation, University of Cape Town, Private Bag X3, Rondebosch 7701, South Africa

[d] Global Science, The Nature Conservancy, Cape Town, 7945, South Africa

Corresponding Author [*]: geethen.singh@gmail.com

**Abstract**

Reliable classification of Earth Observation data depends on consistent, up-to-date reference labels. However, collecting new labelled data at each time step remains expensive and logistically difficult, especially in dynamic or remote ecological systems. As a response to this challenge, we demonstrate that a model with access to reference data solely from time step $t_0$ can perform competitively on both $t_0$ and a future time step $t_1$, outperforming models trained separately on time-specific reference data. This finding suggests that effective temporal generalization can be achieved without requiring manual updates to reference labels beyond the initial time step $t_0$. Drawing on concepts from change detection and semi-supervised learning (SSL), the most performant approach, "Common Ground", uses a semi-supervised framework that leverages temporally stable regions—areas with little to no change in spectral or semantic characteristics between time steps—as a source of implicit supervision for dynamic regions. We evaluate this strategy across multiple classifiers, sensors (Landsat-8, Sentinel-2 satellite multispectral and airborne imaging spectroscopy), and ecological use cases, including local-scale invasive tree species mapping and near-continental-scale land cover mapping. For the invasive tree species mapping, we observed a 21-40% improvement in classification accuracy using Common Ground compared to naïve temporal transfer, where models trained at a single time step are directly applied to a future time step. We also observe a 10 -16% higher accuracy for the introduced approach compared to a gold-standard approach whereby models are trained separately on time-specific reference data. In contrast, when broad land cover categories were mapped across Europe, we observed a more modest 2% increase in accuracy compared to both the naïve and gold-standard approaches. These results underscore the effectiveness of combining stable reference screening with SSL for scalable and label-efficient multi-temporal remote sensing classification.

## 1. Introduction

Remote sensing has become an indispensable tool in contemporary ecology (Pettorelli et al., 2014; Turner et al., 2015) however, its utility depends critically on accuracy and temporal relevance. Among the many machine learning frameworks used to convert spectral data into ecologically meaningful classes, supervised classification remains the most widely adopted, relying on labelled "reference" samples to train models that aim to generalize across space and time (Foody, 2002; Guisan & Thuiller, 2005). The utility of the classification is heavily dependent on the accuracy of the reference labels at the time the imagery was collected (Hauser et al., 2025).

When repeating supervised classification analyses to update data products or look at change over time, a fundamental challenge arises. Reference data collected at the initial time step, $t_0$, may no longer represent conditions at a future time step, $t_1$, due to spectral distribution shifts caused by changing atmospheric conditions, growing vegetation, seasonal or phenological changes and landscape changes such as vegetation succession, disturbances, or land cover transitions (Filippelli et al., 2024; Nagol et al., 2015). If the temporal misalignment between training labels and target conditions are ignored and older reference data naïvely applied at $t_1$, it can cause classification accuracy to decline over time, particularly in dynamic ecosystems (Foody, 2010; Zhu et al., 2017). At the other problematic extreme, collecting new reference data for training a new model at each future time step is logistically challenging, costly, and often infeasible for operational monitoring programs (Gómez et al., 2016), especially in remote or protected areas. Subsequently, two main types of intermediate approaches aim to maintain classification performance while reducing labeling requirements: domain adaptation and filter-based sample migration techniques.

Domain adaptation methods, adapt and transfer learned representations from labeled source domains to unlabeled target domains at different time periods or regions (Tuia et al., 2016; Volpi et al., 2017). However, modern approaches focus on neural network models that are computationally intensive and require large, annotated datasets—constraints that limit their applicability in resource-limited ecological monitoring contexts (Ma et al., 2024). For example, a convolutional neural network trained to classify vegetation condition during a wet year can be fine-tuned with a small number of labeled samples from a subsequent dry year to account for phenological differences, reduced canopy greenness, and increased soil exposure. By updating only the final layers while retaining the spectral–spatial features learned from the wet-year imagery, the model can adapt to inter-annual variability without requiring extensive new annotations. In contrast, filter-based sample migration approaches can function effectively by identifying spectrally stable regions between time points and using them to guide the temporal transfer of class labels (Waldner & Diakogiannis, 2020). Filter-based methods can leverage spectral invariance or change trajectories, making them compatible with operational settings where only pixel subsets remain unchanged and reference data are not continuously updated (Inglada et al., 2017).

Bitemporal filter-based change detection methods that leverage spectral invariance originated from early spectral differencing approaches, including image differencing, image ratioing, and

Change Vector Analysis (CVA) (Lambin & Strahlers, 1994; Malila, 1980; Singh, 1989). CVA pioneered the characterization of both magnitude and direction of spectral change vectors in multivariate space, providing insights into the nature as well as the presence of change. While computationally efficient and requiring minimal prior information, these methods exhibited sensitivity to spectral distribution shifts, noise, and intra-class variability (He et al., 2021). Statistical approaches advanced with Multivariate Alteration Detection (MAD) and its iterative variant IRMAD, based on canonical correlation analysis (Nielsen, 2007). IRMAD iteratively applies canonical correlation analysis between image pairs, down-weighting pixels contributing to spectral change while emphasizing stable areas. This spectral domain approach operates without requiring access to labelled data.

Trajectory-based methods model spectral evolution across dense EO time series to detect and classify land cover changes. Continuous Change Detection and Classification (CCDC) employs harmonic regression to identify statistically significant change points in pixel-wise reflectance time series (Z. Zhu & Woodcock, 2014). These breakpoints partition time series into spectrally stable segments that are subsequently classified using supervised models. Classification operates at the segment level using features derived from temporal coefficients rather than raw reflectance, reducing the need for dense labeling by allowing single labels to describe entire temporally stable segments (Zhu et al., 2019). While the scalability and effectiveness of using stable regions based on trajectory based methods are supported by numerous multi-temporal national to global mapping initiatives (Zhang et al., 2023; Zhou et al., 2025), these techniques can be constrained by cloud cover, sensor or processing differences, and limited revisit frequency—especially in experimental or newly launched missions (Zhu et al., 2022). They are also prone to false positives in dynamic ecosystems prone to natural disturbances such as fire or drought (Zhu et al., 2022; Z. Zhu & Woodcock, 2014).

In contrast to prior sample migration approaches that rely on bi-temporal or multi-temporal change detection to identify and discard samples affected by change, we instead retain both the initial $t_0$ reference samples (with $t_0$ imagery) and the subset of these samples that remain spectrally and structurally stable at $t_1$ based on $t_1$ imagery. These jointly constitute the labelled set, which is then used to provide pseudo-labels for changed samples through a semi-supervised learning (SSL) framework. Our SSL approach, referred to as "Common Ground", is motivated by practical needs. Ecologists and conservation practitioners increasingly rely on maps that are updated over time, often with access to limited or outdated field data (Hauser et al., 2016), and/or based on imagery from different sensors over time. This is particularly true in small-area mapping tasks, such as monitoring invasive species or habitat condition within protected areas, where national and global landcover datasets lack the specificity and accuracy needed for fine-scale dynamics (Gillespie et al., 2008). Moreover, computationally intensive solutions like neural-network-based domain adaptation (Zhu et al., 2021) are often impractical in such settings due to resource constraints or poor representation of locally relevant land cover classes at a broad enough extent to allow the collection of sufficient reference data. Our proposed method is lightweight, compatible with both traditional classifiers (e.g., Random Forests) and modern models (e.g., transformer-based architectures) and does not require relabelling or dense time series, thus avoiding the technical, software, and hardware overhead required for larger-scale mapping that exceeds local jurisdiction needs and budgets.

The primary objective of this study is to evaluate "Common Ground", a semi-automated framework, for improving temporal transferability of reference data used to train land cover classifiers by explicitly addressing label degradation and spectral distribution shifts. Specific aims include: (1) assessing how the Common Ground approach compares to alternate strategies for selecting and reusing reference labels in terms of classification performance; (2) evaluating performance across classifier types (traditional Random Forest and modern transformer-based tabular foundation model, TabPFN v2) and remote sensing modalities, including Landsat-8 and Sentinel-2 multispectral and airborne imaging spectroscopy (hyperspectral imagery); and (3) benchmarking the framework in two ecological use cases: broad-class land cover classification at a near-continental scale and fine-class invasive species detection at a local landscape scale. By testing across multiple model architectures, sensors, and ecological targets, this study assesses whether the Common Ground approach demonstrates broad applicability for operational monitoring programs.

## 2. Methods

We compare five strategies for the temporal transfer of reference labels (Table 1), ranging from naïve reuse to semi-supervised learning with spectral stability constraints. To assess their robustness, we applied these strategies across three diverse case studies where reference labels were available for two distinct time steps. Refer to Table A. 1. for details on training data allocations. While we utilise reference labels for the target time step ($t_1$) to validate performance, the primary objective is to demonstrate accuracy for practical applications where $t_1$ labels are unavailable. The first two case studies used the same reference labels to perform fine-grained invasive tree species detection in South Africa: one using multispectral imagery from Sentinel-2, and another using airborne imaging spectroscopy (hyperspectral imagery). The airborne data were collected using different imaging spectrometers for two timepoints, allowing this case study to include a test of cross-sensor transferability. The third case study explored near-continental land cover mapping across a large subset of Europe. Using the large reference dataset in Case Study 3, we explicitly evaluate the effect of training sample size on the performance of Common Ground. Each case study varied systematically in spatial extent, sensor configuration, class complexity, sample size and temporal variability, enabling evaluation of different strategies under varying operational constraints.

### 2.1.1. Experimental Design

**Table 1.** Overview of approaches evaluated for generating a $t_1$ classification map. Experiments differ mainly in training data composition, reflecting varying degrees of manual relabelling, use of change masks, and pseudo-labelling. The gold standard (Experiment 1) uses exhaustive $t_1$ relabelling. Experiment 2 assumes temporal stability and applies a $t_0$-trained model to $t_1$ imagery. Experiments 3–5 reduce manual effort via spectral stability filtering and semi-supervised learning: Experiment 3 samples pseudo-labels from stable areas, Experiment 4 combines $t_0$ labels and imagery with the stable subset of $t_0$ labels and $t_1$ imagery, and Experiment 5 generates pseudo-labels for changed areas using the Experiment 4 model.

| Experiment | Training Data for $t_1$ model | IRMAD Used | Pseudo-Labels | Manual Relabelling | Description |
|---|---|---|---|---|---|
| **Exp. 1** Gold Standard | Fully relabelled $t_0$ labels at $t_1$ with $t_1$ imagery | ✗ | ✗ | ✓ | Manual review and reclassification of all $t_0$ labels at $t_1$. High cost, not scalable. |
| **Exp. 2** Naïve Transfer | $t_0$ imagery and $t_0$ labels | ✗ | ✗ | ✗ | Directly applies a model trained on $t_0$ imagery and labels to $t_1$ imagery. Assumes stationarity. |
| **Exp. 3** Spectral Stability Sampling (Wessels et al., 2016) | Pseudo-labels sampled from $t_0$ classification map in unchanged areas | ✓ | ✓ | ✗ | IRMAD filters spectrally stable regions. $t_0$ model-derived (pseudo) stable labels used as training data. |
| **Exp. 4** Cross-temporal Stable Points | $t_0$ labels and imagery + stable $t_1$ labels and imagery | ✓ | ✗ | ✗ | Combines $t_0$ labels in spectrally stable regions with $t_0$ and $t_1$ imagery. No label generation. |

| Exp. 5 Semi-Supervised Learning (Common Ground) | $t_0$ labels and imagery + stable $t_1$ labels and imagery + pseudo-labelled $t_1$ labels and imagery for change subset | ✓ | ✓ | ✗ | Changed $t_1$ subset pseudo-labelled using (exp 4) model trained on all $t_0$ labels and stable $t_1$ labels. |
|---|---|---|---|---|---|

### 2.1.2. Detailed experimental protocols

- **Experiment 1 – Fully Relabelled Supervised Learning (Gold Standard)**: To serve as an expected upper-bound benchmark, a separate model is trained using manually updated labels at $t_1$. Each reference point from $t_0$ is visually reviewed against $t_1$ imagery, reclassified where necessary, and used to train a new model on $t_1$ imagery. While this exhaustive approach can adapt to label changes and spectral shifts, it is labour-intensive and not scalable to large or frequently updated classification datasets, motivating this study.
- **Experiment 2 – Naïve Temporal Transfer Baseline**: A classifier is trained on reference labels and imagery data from an initial time step ($t_0$) and directly applied to later imagery ($t_1$) (Experiment 2.1). This approach assumes temporal stationarity in both label validity and spectral features—a condition often violated in dynamic landscapes. To partially address spectral shifts, L2 normalization is applied to harmonize reflectance distributions across time (Experiment 2.2). Although simple, L2 normalization does not address potential changes in land cover class labels between $t_0$ and $t_1$.
- **Experiment 3 – Spectral Stability Sampling (Wessels et al., 2016):** This approach follows the method introduced by Wessels et al. (2016), in which a binary change/no-change mask is first generated for the study area. Binary change masks are derived using the Iteratively Reweighted Multivariate Alteration Detection (IRMAD) algorithm for the invasive species case studies and the Continuous Change Detection (CCD) algorithm for the LUCAS case study, enabling fully automated identification of changed and unchanged areas between $t_0$ and $t_1$. Rather than relying on pre-existing reference labels, new reference samples are drawn from the $t_0$ classification map, but only within areas identified as stable between $t_0$ and $t_1$, effectively treating the map itself as a source of pseudo-labels to train a new model. The core assumption is that spectrally unchanged regions retain their land cover class, allowing the $t_0$ classification to serve as a proxy for $t_1$ labels. While more time-efficient than full manual relabelling, this method is susceptible to error propagation from the $t_0$ classification map and may require large sample sizes to average out noise. It is particularly ill-suited to small-area or fine-grained mapping tasks, where class imbalance and localized errors in the $t_0$ map can have outsized impacts on model performance (Foody, 2010). Finally, this approach is less suitable for experimental or research-focused applications due to the requirement of generating an intermediate $t_0$ map and the potential for bias propagation arising from area-proportional sampling based on pixel counts (Olofsson et al., 2014; Stehman & Czaplewski, 1998; Stehman & Foody, 2019).

- **Experiment 4 – Naïve Baseline with Cross Temporal Training and Temporal Stability Constraints:** We combine all $t_0$ reference samples with spectrally stable $t_1$ reference points that have been identified through binary change/no-change assessment via visual assessment (Experiment 4.1) or the IRMAD/CCD algorithm (Experiment 4.2). Changed $t_1$ samples are excluded entirely, and no pseudo-labelling is performed. This approach isolates the effect of incorporating temporally stable $t_1$ reference data without requiring full manual relabelling (Experiment 1) or pseudo-labelling (Experiment 3; (Wessels et al., 2016). Unlike Wessels et al. (2016), who generated new labels from the $t_0$ model-derived map within stable areas, we filtered pre-existing reference data for temporal validity. This experiment evaluates whether binary change filtering of existing labels alone improves temporal generalization without the complexity of class-level label inference, establishing a foundation for Experiment 5's automated pseudo-labelling of changed areas.

- **Experiment 5 – Semi-Supervised Learning with Temporal Stability Constraints (Common Ground):** Extending the previous approach, we investigate a two-stage SSL framework that improves label coverage while maintaining reliance solely on binary change/no-change masks. Unlike Wessels et al. (2016), who derive training labels from the $t_0$ classification map in spectrally stable areas, our method filters existing annotated data and avoids generating new training labels directly from predicted maps.

    · Stage 1: A model is trained using the full set of $t_0$ reference samples, along with $t_1$ reference samples identified as spectrally stable via change detection (the model from the previous experiment (4.1 and 4.2). This model is used to infer pseudo-labels for reference samples marked as changed at $t_1$.

    · Stage 2: The pseudo-labelled $t_1$ samples (i.e., dynamic areas), stable $t_1$ samples, and all $t_0$ samples are combined to train a final model.

This SSL approach provides a scalable and label-efficient alternative to full relabelling. By focusing on existing reference data rather than sampling new data from maps, it reduces the risk of propagating errors while also having lower computational requirements. It also explicitly incorporates dynamic areas into the training process, unlike previous stability-based methods that exclude them entirely.

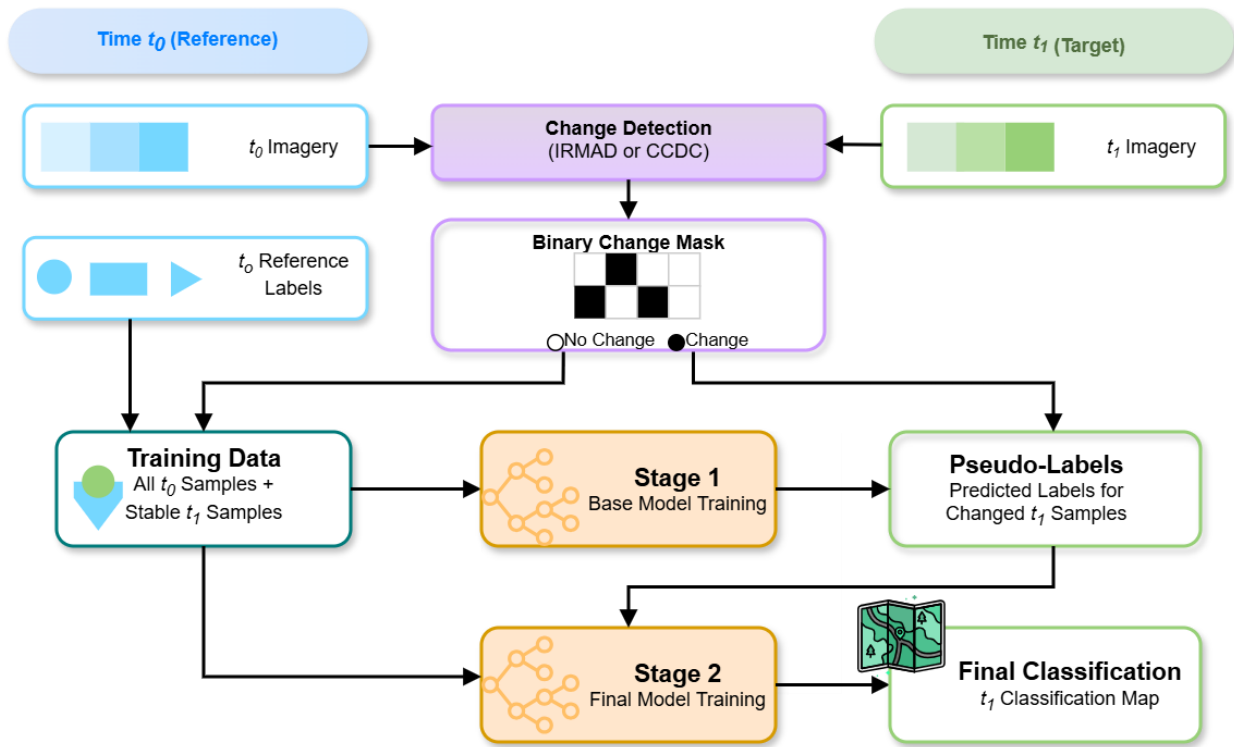

**Fig. 1** The Common Ground workflow, using semi-supervised learning for multi-temporal invasive species and land cover classification (Table 1, experiment 5.2.). Closed boxes with outlines correspond to processing steps, while open boxes with outlines correspond to data inputs or outputs. The two-stage framework combines reference data from time $t_0$ (blue) with a subset of temporally stable $t_0$ reference labels in $t_1$ based on unsupervised change detection between the imagery from the two time steps to train an initial model (Stage 1, orange), which then generates pseudo-labels for spectrally changed areas. The final model (Stage 2, orange) incorporates all $t_0$ data, stable $t_1$ data and pseudo-labeled $t_1$ change samples from dynamic regions, enabling label-efficient adaptation to temporal changes without requiring manual relabeling. Note, the stage 1 model also corresponds to the model used in experiment 4. Refer to the supplementary material, Figure A.1 for experiment workflows 1-4.

## 2.2. Model training and evaluation

For each of the five experimental strategies, we evaluated Random Forest across all experiments, while TabPFN (v2) was applied only to experiments with fewer than 10,000 samples. We selected TabPFN (v2) since it is a modern model with state-of-the-art accuracy on tabular datasets at the time of performing the experiments (Hollmann et al., 2025). Random Forest is a non-parametric ensemble learning algorithm that constructs multiple decision trees during training and outputs the mode of the classes for classification tasks. Their robustness to overfitting, tolerance to noisy features, and interpretability have made them a standard in remote sensing applications (Belgiu & Drăguţ, 2016). TabPFN (v2), on the other hand, is a transformer-based probabilistic model designed specifically for tabular data and few-shot learning settings. It frames classification as a Bayesian inference problem and leverages pretraining on millions of synthetic tasks to make predictions in a single forward pass without fine-tuning (Hollmann et al., 2025). Its ability to generalize from small, labelled datasets makes it well-suited for sparse data settings and handling high-dimensional spectral inputs effectively.

Training and evaluation were conducted using a five-fold spatiotemporal leave-location-and-time-out (LLTO) cross-validation scheme. To construct spatially and temporally independent folds, we applied k-means clustering to the spatial coordinates of all labelled points and assigned folds based on the resulting clusters, ensuring that each fold contained samples from both $t_0$ and $t_1$. In each iteration, the model was trained on specific location-time combinations (e.g., folds A-D at both $t_0$ and $t_1$) and tested on held-out spatial locations and time periods during training (e.g., fold E at $t_1$, while fold E at $t_0$ was excluded from both training and testing), preventing information leakage from temporally stable points and forcing the model to extrapolate simultaneously in both spatial and temporal dimensions. This design controls for both spatial autocorrelation and temporal dependence between training and test sets, providing a stringent assessment of the model's ability to generalize to novel space-time combinations (Meyer et al., 2018; Roberts, 2017). LLTO validation is particularly relevant for operational land cover mapping scenarios where predictions must be made at new locations and time periods not represented in the training data (Ploton, 2020).

For Experiment 3 (Wessels et al., 2016), validation was instead carried out on the full set of available reference points for 2023. Because Experiment 3 does not use these reference points for training (training uses stratified random samples drawn from the stable areas of the 2018 model-derived classification map), we removed any validation points within 100 m of a training sample to reduce spatial dependence. In all experiments, model performance was assessed using mean F1 score and overall accuracy across folds. Standard deviations for each metric were also reported to assess variability in performance. The macro F1 score was emphasized as the primary metric due to its robustness to class imbalance and relevance for rare invasive species detection.

### 2.3. Inference

For each experiment, models were trained using all data available under the respective training configuration. Inference was conducted on randomly selected image patches using the fully trained models. For TabPFN v2, inference and evaluation were performed only when the training set size and the number of pixels within a patch selected for inference were within the model's supported capacity (≤10,000 samples, (Hollmann et al., 2025); configurations exceeding this limit were therefore excluded from TabPFN-based evaluation.

### 2.4. Case studies

The first case study evaluates the detection and classification of three genera of invasive trees under operational constraints typical for conservation authorities. Specifically, limited labelled data (< 1 600 labelled instances), fine-grained class structure with a locally-relevant targeted thematic interest and freely available imagery (Sentinel-2 multispectral). This tests the SSL framework under conditions of high spatial heterogeneity and limited data.

Using the same geographic extent, the second case study examines temporal transfer across airborne hyperspectral sensors, reflecting real-world monitoring programs that must adapt to new and evolving satellite missions and/or opportunistically harness available datasets. The third case study expands to a near-continental scale, applying the framework to land cover mapping using LUCAS ground survey data and bi-temporal Landsat imagery to test transferability across years, biomes, and multiple multispectral sensors.

### 2.4.1. Case study 1 and 2: Invasive alien trees in protected areas of the Western Cape, South Africa

This case study examines invasive alien trees within protected areas of the Western Cape, South Africa, including parts of the Cape Floral Region Protected Areas UNESCO World Heritage Site. The Cape Floristic Region is the smallest yet most diverse floral kingdom in the world, harbouring over 9 000 plant species, of which approximately 69 % are endemic (Goldblatt & Manning, 2002) In this biodiversity hotspot, invasive alien plants are recognised as the single greatest threat to native biodiversity and ecosystem functioning (Wilson et al., 2014).

Three dominant invasive tree genera—*Acacia*, *Eucalyptus*, and *Pinus*, commonly known as wattles, gums and pines respectively—were selected due to their high ecological impact, extensive distribution, and significant reduction of water availability through increased evapotranspiration (Le Maitre et al., 2016). The focal study area encompasses the upper Berg and Breede River catchments, which receive ~380 mm mean annual rainfall and are classified as water-scarce. Invasive stands in these catchments have been shown to substantially reduce streamflow and groundwater recharge (Le Maitre et al., 2016; Scott & Prinsloo, 2008), making early detection and monitoring essential for biodiversity conservation and water resource management.

#### 2.4.1.1. Reference data preparation

Reference data for invasive alien tree species were generated using a combination of high-resolution Google Earth imagery and expert knowledge of local species distributions in the Western Cape, South Africa. Initial (2018) reference points were manually digitized to represent both invaded and surrounding land cover types (Figure 2). To reflect landscape changes between 2018 and 2023, all reference points were systematically reviewed using multi-source visual interpretation. This process involved cross-referencing Very High Resolution (VHR) imagery from Google Earth, South Africa's National Geo-spatial Information (NGI) mapping agency, and ESRI's World Imagery basemap. Each reference point was first assigned a binary label indicating "change" or "no change." For all locations identified as changed, a new land cover class label (i.e., "to-class") was assigned. In cases where changes were minimal and did not meet the threshold for a class transition (e.g., minor reduction in vegetation density in fynbos), the binary label was reverted to "no change" to ensure intra-class consistency and reduce label noise. If the to-class could not be assigned confidently, a label was not assigned for 2023 (n = 41 points) and not used during training.

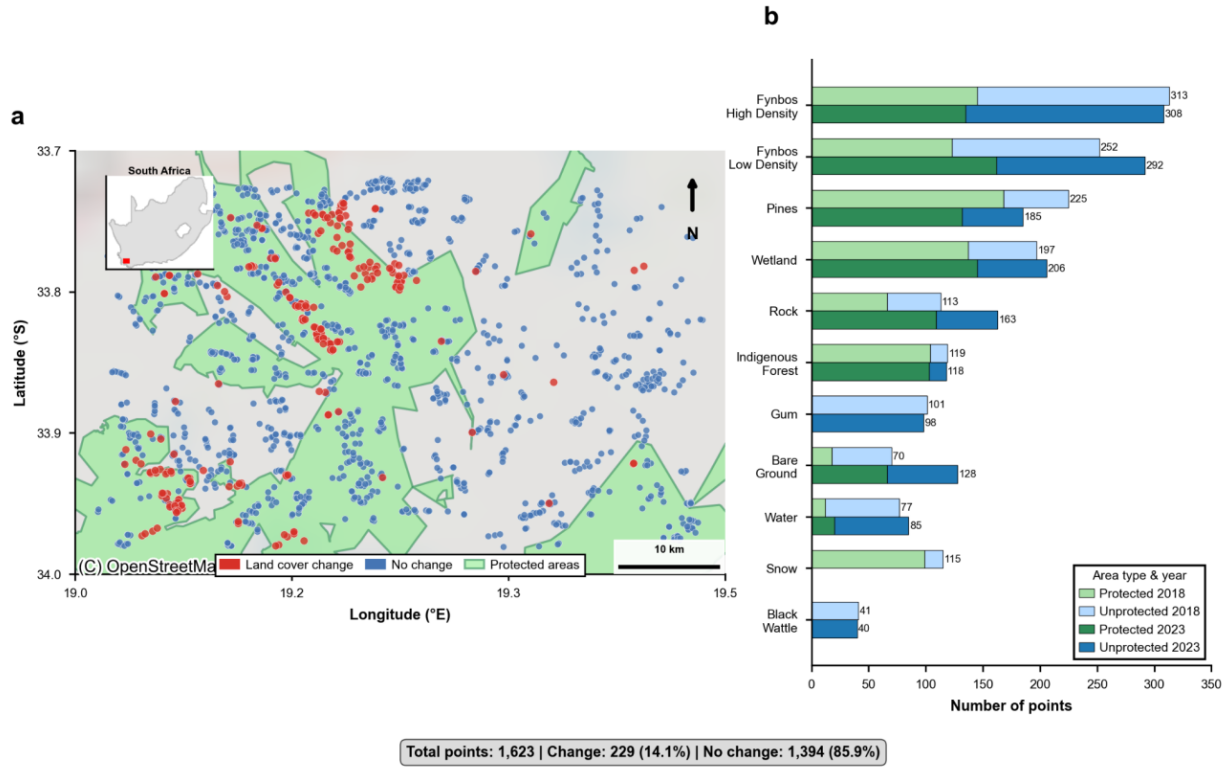

**Fig. 2** Changes in land cover distribution between protected and unprotected areas from 2018 to 2023, based on reference samples. (a) Map of the study area in The Western Cape showing the spatial distribution of reference points used for invasive tree species classification. Points are colored by their change status. The Cape Floristic Region protected area boundaries are overlaid in green. (b) Stacked horizontal bar chart showing the number of reference points for each land cover class, stratified by protection status (green: protected; blue: unprotected) and year (lighter shades: 2018; darker shades: 2023). The total number of points per class is annotated at the end of each bar. The distribution highlights shifts in land cover composition, with notable declines in invasive tree species and increases in bare ground and low-density fynbos over the five-year period.

### 2.4.1.2. Sentinel-2 composite generation

We utilized the Sentinel-2 Level-1C top-of-atmosphere (TOA) imagery available on GEE to construct cloud-free composites for two time points: September 2018 and November 2023. These periods were selected to match the availability of the hyperspectral data. For each period, we filtered the image collection to the respective months and applied the *Cloud Score Plus* algorithm for cloud and shadow masking. Following masking, we excluded atmospheric bands (bands 1, 9, and 10) and applied L2-norm brightness normalization across the remaining spectral bands to account for variation in illumination and sensor angle. The resulting composites represent spectrally normalized surface conditions during the late dry season for each year.

### 2.4.1.3. Hyperspectral imagery processing

The September 2018 hyperspectral data were collected using a Fenix 1K sensor that captures data at 6-nm intervals across the 390–2450 nm spectral range (Visible to Shortwave Infrared, VSWIR), yielding a total of 322 bands at a 5m spatial resolution. The data used is provided as a level-1B surface reflectance product that has been atmospherically corrected based on ATCOR4, geometrically corrected and radiometrically balanced to compensate for Bi-Directional Reflectance Functions (BDRF). The 2023 hyperspectral data was collected using NASA's AVIRIS-NG sensor as part of the Biodiversity Survey of the Cape (BioSCape) campaign (Cardoso et al., 2025). The AVIRIS-NG instrument is a high signal-to-noise ratio (SNR) pushbroom sensor that captures data at 5-nm intervals across the 380–2510 nm spectral range (Visible to Shortwave Infrared, VSWIR), yielding a total of 425 bands at a 5m spatial resolution. The 2023 acquisition occurred during 21 October to 25 November over the study area (Figure 2).

We used the publicly released Level-3 (L3) surface reflectance mosaic v2 product for the 2023 AVIRIS-NG data (Brodrick et al., 2025). These data were pre-processed by NASA and included atmospheric correction, spatial resampling to 5-m resolution, and tiling into a regular grid of 807 tiles. Each tile integrates multiple scenes from overlapping flight lines acquired over several days. For our analysis, we removed atmospheric absorption bands. To enable direct spectral comparison between datasets collected by sensors with differing band centers, we implemented a Gaussian-based band resampling procedure. Reflectance spectra from the 2018 sensor were resampled to match the band configuration of AVIRIS-NG, which offers approximately 5 nm full width at half maximum (FWHM) spectral resolution across the 380–2510 nm range. The resampling process integrated each source band over the Gaussian-shaped response profiles of the AVIRIS-NG target bands wherever spectral overlap occurred. Bands in the target sensor that lacked any overlap with the source bands were assigned NaN values (n = 131 bands, 30.8 **%** of total target bands), reflecting the absence of spectral support. This method yielded resampled spectra that are spectrally aligned and suitable for direct comparison across sensors, minimizing structural bias due to differences in sensor-specific spectral discretization.

### 2.4.1.4. Change detection and spectral stability assessment

The IRMAD algorithm (Canty & Nielsen, 2012) was used to identify spectrally stable pixels between the initial ($t_0$) and second ($t_1$) Sentinel-2 composite in Experiments 3, 4.2, and 5.2. To derive a binary change mask, we empirically determined thresholds by comparing IRMAD outputs to known examples of change and stability. Initial trials using the recommended 99.9th percentile threshold (Canty & Nielsen, 2012) and the minimum-error Bayes discriminant method (Wessels et al., 2016) produced unsatisfactory results in our context. Given the pronounced class imbalance between changed and unchanged pixels, we instead selected thresholds based on precision–recall curve analysis, which is better suited for imbalanced datasets. This yielded a threshold of 95.811. The Sentinel-2-derived mask was used for both invasive tree species case studies.

Following Wessels et al. (2016), spectrally stable areas (pixels below the change threshold) were used to mask the $t_0$ classification and $t_1$ spectral data prior to sampling. Similar to Wessels et al. (2016), we aimed to sample at least 0.01 % of the study areas stable areas (160 708 pixels). Subsequently, we calculated class-specific sample sizes using an area-weighted allocation scheme (Colditz, 2015; Wessels et al., 2016). To ensure adequate representation, a minimum of 1 608 samples was assigned per class when the area-based calculation yielded a smaller number, producing 162 294 samples for training a model for $t_1$ (Experiment 3).

### 2.4.2. Case study 3: Broad-class land cover classification across a large subset of Europe

This case study utilised the Land Use/Cover Area frame statistical Survey (LUCAS), a harmonised, in situ dataset collected by Eurostat to support consistent land cover and land use monitoring across Europe. LUCAS provides field-verified point data on land use, land cover, and environmental parameters at a continental scale, surveyed on a regular three-year cycle since 2006. The dataset covers all EU Member States using a systematic grid-based sampling design, thereby ensuring statistical representativeness across diverse European biomes and land cover types, making it ideal for testing the scalability and generalizability of our Common Ground approach across large spatial extents and varied environmental conditions. LUCAS observations serve both as ground-truth information for validating remote sensing products and as an independent reference for large-scale land cover classification, calibration of Earth observation algorithms, and long-term monitoring of land-use change dynamics (d'Andrimont et al., 2020; Eurostat, 2020). In this study, LUCAS data were employed to calibrate and evaluate classification accuracy and to provide a robust empirical evaluation of Common Ground for large datasets and large-area mapping.

### 2.4.2.1. Reference data preparation

We utilized LUCAS survey data from 2015 and 2018 to match the availability of Landsat-8 imagery. While the LUCAS dataset contains comprehensive land cover classifications with up to 82 land cover classes, we aggregated these into eight broader categories used in a prior LUCAS-based land cover map (ELC10) to ensure sufficient sample sizes per class. The 2018 reference dataset prepared by the ELC10 authors was used (Venter & Sydenham, 2021), as it underwent two cleaning and filtering steps to improve reliability (Pflugmacher et al., 2019; Weigand et al., 2020). Firstly, heterogeneous points with less than 50% parcel cover were excluded, followed by the removal of thematically and spectrally ambiguous classes, including "Linear features" (LC1 code A22), "Other artificial areas" (A39), "Temporary grasslands" (B55), "Spontaneously re-vegetated surfaces" (E30), and "Other bare land" (F40). Secondly, an outlier ranking procedure based on Sentinel-2 spectra and random forest classification confidence was applied to identify archetypal versus contaminated points (Venter & Sydenham, 2021), resulting in the retention of 18,009 high-confidence samples that supplemented the LUCAS polygon centroid dataset (71 485). For our study, we further restricted the dataset to only those 2018 points that had a corresponding 2015 observation, resulting in 33 734 points for each year, enabling consistent temporal comparison. This curated continental-scale dataset enables assessment of temporal transfer performance across different climate zones, land cover types and land use intensities, and biogeographic regions within Europe, providing a robust test of method scalability beyond the fine-grained, localized scenarios represented by the invasive species case studies.

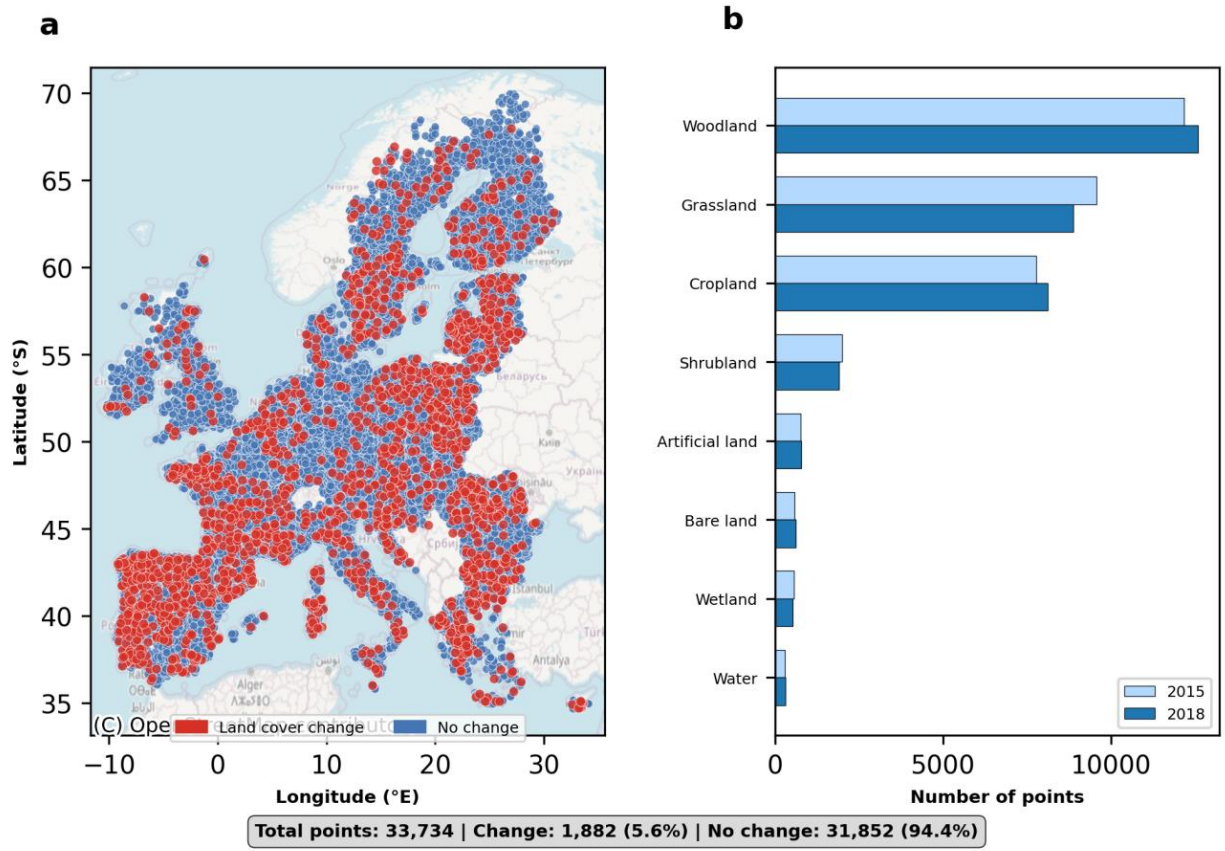

**Fig 3.** Changes in land cover distribution between 2015 and 2018, based on LUCAS reference samples. (a) Map of the European study area showing the spatial distribution of LUCAS points used for land cover classification. Points are colored by their change status. (b) Horizontal bar chart showing the number of reference points in each land cover class for 2015 (light blue) and 2018 (dark blue). The total number of points per class is annotated at the end of each bar. The distribution illustrates shifts in land cover composition over the three-year period, including reductions in grassland and increases in woodland and cropland.

#### 2.4.2.2. Landsat-8 composite generation

We utilized the Landsat-8 surface reflectance imagery available on GEE to construct cloud-free composites for the two latest survey years (2015 and 2018). These periods were selected to match the availability of Landsat-8 data, only available post-2013. For each year, we created an annual composite by filtering the Landsat-8 image collection to the required year and applied cloud and cloud shadow masking using the accompanying quality bands. Following masking, we excluded atmospheric, panchromatic, and thermal bands (bands 1, 8-11) and applied L2-norm brightness normalization across the remaining spectral bands to account for variation in illumination and sensor angle. Finally, we created a median image composite from all remaining pixels. The resulting composites represent the median spectrally normalized surface conditions for each year.

### 2.4.2.3. Change detection and spectral stability assessment

For the much larger land cover case study, we used the readily computed CCD dataset available on Google Earth Engine (GEE) to obtain a binary change mask between 2015 and 2018 that was used in experiments 3, 4.2 and 5.2 (ee.ImageCollection("GOOGLE/GLOBAL_ CCDC/V1")). For experiment 3, we applied this change layer to a 2015 Landsat-derived landcover layer presented in Pflaugmacher et al. (2019) that was reclassified to match our 8-class scheme (Figure 3). Following the masking of stable areas, we performed stratified random sampling to obtain 10 000 points across each of the 8 classes that were used for training a model for $t_1$ during experiment 3.

### 2.5. Implementation details

All experiments were implemented in Python 3.12. The Random Forest classifier was used from the scikit-learn package (version 1.4.2; Pedregosa et al., 2011). TabPFN v2 was obtained from the official [TabPFN GitHub repository](#) associated with Hollmann et al. (2025), using the most recent commit at the time of writing (commit c2d83e0, April 2024). Band resampling was achieved using the BandResampler module from the Spectral Python (SPy) library. Sentinel-2 multispectral imagery was accessed and processed via the Google Earth Engine Python API (version 0.1.381, Gorelick et al., 2017)). The IRMAD change detection was implemented using the Python scripts developed by Mort Canty (Canty, 2025). Visualization and figure generation were performed using Matplotlib (version 3.8.4) and Seaborn (version 0.13.2).

### 3. Results

### 3.1. Case study 1 and 2: Invasive Species (Multispectral and Hyperspectral)

We evaluated invasive tree species classification using both multispectral (Sentinel-2) and hyperspectral (airborne) imagery across five experimental approaches, employing Random Forest (RF) and TabPFN v2 classifiers (Table 2). Overall, our Common Ground approach, combining iteratively reweighted multivariate alteration detection (IRMAD) with semi-supervised learning (SSL) yielded the best results across both sensor types and classifiers. This two-stage approach (Exp. 5.2) consistently achieved the highest mean F1-scores, with improvements of up to 0.15 compared to baseline models. Importantly, IRMAD proved sufficient both quantitatively and qualitatively to automate the labelling of binary change/no-change subsets (Table 2; Figure 4), eliminating the need for manual relabelling and facilitating scalable training data generation.

Across classifiers, TabPFN typically outperformed RF by up to 0.15, particularly under high-dimensional hyperspectral approaches. This suggests that TabPFN is better suited for high-dimensional spectral data. However, multispectral data performed competitively when combined with SSL, reducing the performance gap relative to hyperspectral input. That said, the performance of the hyperspectral analysis may have been better if the labelling of binary change/no-change subsets were based on hyperspectral data, as opposed to relying on the Sentinel data as it did here. Without SSL, hyperspectral data tended to outperform multispectral imagery, especially for TabPFN, reinforcing the benefit of SSL in bridging sensor and dimensionality differences.

When different sensors were used for training and testing, applying L2 normalisation substantially improved performance. In some cases (Exp. 4.1–4.2), normalisation boosted F1-scores by more than 30%. Further improvements were observed when combining normalisation with cross-temporal training on a stable subset of spectra, whether manually labelled or derived from IRMAD. Under these conditions, both RF and TabPFN (with hyperspectral data) achieved competitive performance relative to same-sensor multispectral approaches.

The Wessels et al. (2016) approach offered modest improvements over a naïve baseline with L2 normalisation but did not outperform any of the other tested approaches, including the gold standard or naïve cross-temporal training strategies. This highlights the importance of combining spectral normalisation with temporal stability information for robust cross-sensor applications.

Finally, the results show that SSL is most effective when coupled with IRMAD (Exp. 5.2), where multispectral and hyperspectral inputs both achieved their highest F1-scores. This demonstrates the critical role of unsupervised change detection in supporting semi-supervised model training for the invasive tree species mapping case study.

**Table 2.** Mean (standard deviation) F1-scores after 5-fold spatial cross-validation for invasive tree species classification. Results are shown for multispectral (Sentinel-2, Case study 1) and hyperspectral (airborne, Case study 2) imagery, using Random Forest (RF) and TabPFN v2. Bold values indicate the best-performing approach per case study and per model. For Experiment 3 (Wessels et al., 2016), performance was evaluated once on the full filtered validation set (rather than across the 5-fold LLTO splits), and therefore no standard deviation is reported. We also do not report TabPFN results for Experiment 3 because it does not support training sets larger than 10,000 samples.

|  | Exp. ID | Satellite Multispectral (Sentinel-2) |  | Airborne Hyperspectral |  |
|---|---|---|---|---|---|
|  |  | *Random Forest (RF)* | *TabPFN v2* | *Random Forest (RF)* | *TabPFN v2* |
| **Gold standard** | 1 | **2018:** 0.52 (0.07) | **2018:** 0.64 (0.04) | **2018:** 0.59 (0.03) | **2018:** 0.74 (0.07) |
|  |  | **2023:** 0.51 (0.04) | **2023:** 0.57 (0.03) | **2023:** .49 (0.06) | **2023:** 0.62 (0.09) |
| **Naïve Baseline** | 2.1 | 0.43 (0.03) | 0.50 (0.06) | 0.01 (0.01) | 0.01 (0.01) |
| + Normalisation | 2.2 | 0.43 (0.03) | 0.48 (0.04) | 0.36 (0.06) | 0.32 (0.08) |
| Wessels et al. (2016) | 3 | 0.49 | - | 0.46 | - |
| Naïve Baseline + Normalisation + Manually labelled stable subset | 4.1 | 0.56 (.05) | 0.62 (.05) | 0.57 (.05) | 0.70 (0.07) |
| Naïve Baseline + Normalisation + IRMAD stable subset | 4.2 | 0.54 (0.05) | 0.62 (0.05) | 0.58 (0.02) | 0.71 (0.06) |
| Manual relabelling (stage 1) + SSL (stage 2) | 5.1 | 0.58 (.04) | 0.67 (0.04) | 0.57 (0.04) | 0.71 (0.07) |
| **IRMAD (stage 1) + SSL (stage 2, Common Ground)** | 5.2 | **0.64 (.03)** | **0.73 (.03)** | **0.64 (.03)** | **0.72 (.05)** |

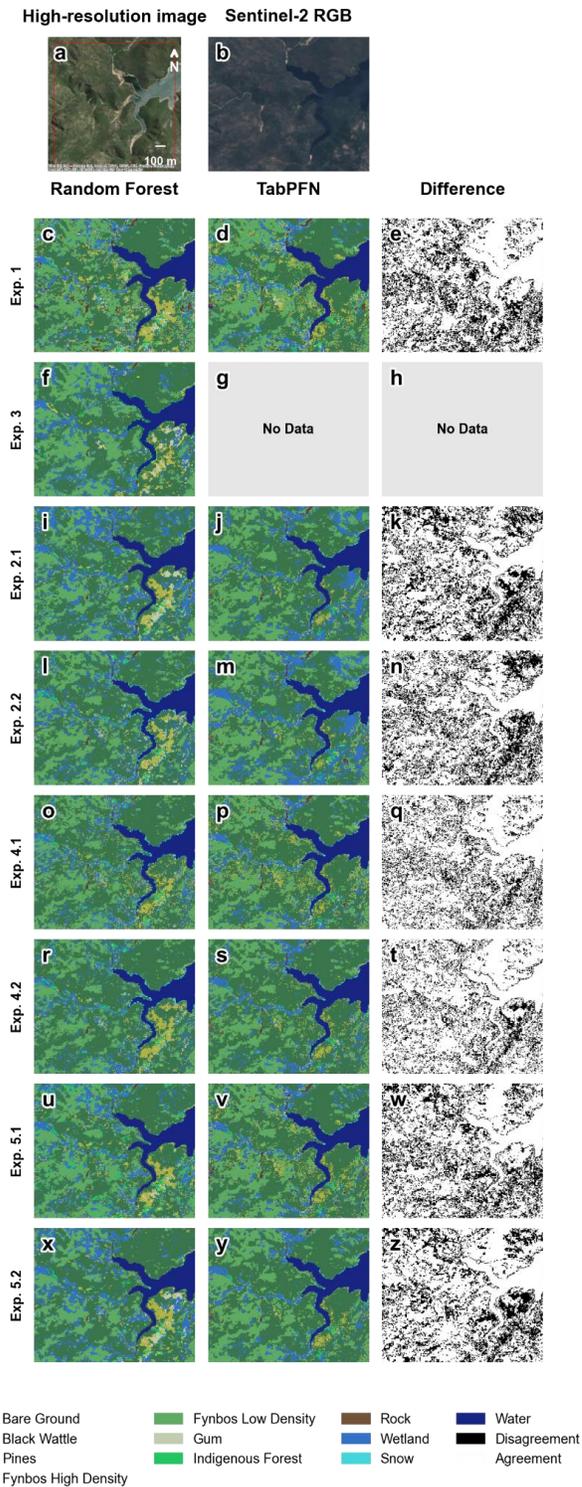

**Fig. 4.** Qualitative comparison of predictions across the five experimental approaches for the multispectral invasive species case study. Panels show: (a) high-resolution ESRI imagery, (b) RGB Sentinel-2 imagery from 2023, and predictions from two models (Random Forest and TabPFN in columns 1 and 2, respectively) and their (dis)agreement (column 3). Specifically, (c-

e) the gold-standard requiring complete relabelling, (f-h) Wessels et al. (2016), (i-k) naïve baseline, (l-n) naïve baseline with normalisation, (o-q) cross-temporal training with spectrally stable data identified using manually labelled change, (r-t) cross-temporal training with spectrally stable data identified using unsupervised change detection, (u-w) semi-supervised learning using manually labelled change, and (x-z) semi-supervised learning using unsupervised change detection. Note, there are no predictions for Exp. 3 (g-h), TabPFN since the number of pixels exceeded 10 000 and the subsequent prohibitively slow inference speeds. See A.2 and S3 for more examples.

### 3.2. Case study 3: LUCAS land cover mapping (Multispectral)

Consistent with findings from Case Studies 1 and 2, the best-performance was obtained using SSL with CCDC-derived change/no-change labelling (Exp. 5.2), which achieved the highest mean F1-score (0.71). For the larger-scale LUCAS land cover classification experiment (>33,000 samples, Landsat-8) across a near-continental extent, the SSL framework continued to yield performance improvements relative to time-specific supervised models (Table 3). However, the gains were more modest than those observed in the invasive species case studies with smaller training sets (~1,500 samples). Specifically, while SSL improved mean F1-scores by up to 0.02 (from .69 to .71) compared to the gold standard time-specific model training, this represents only a 2-3% relative gain compared to ~16-30% relative gains in the invasive species setting, indicating diminishing returns of SSL depending on mapping targets as opposed to the abundance of reference data available (Figure 5).

Among stability information sources, both manually labelled stable subsets and CCD-derived subsets proved effective (Exp. 4.1–4.2), producing results similar to or slightly better than the gold-standard supervised baseline. This highlights CCDC as a viable, automated alternative to manual labelling for generating stable subsets.

Applying L2 normalisation had a detrimental effect on performance in this case, reducing F1-scores by approximately 9% (Exp. 2.2). This contrasts with the invasive species case study using hyperspectral data, where normalisation improved cross-sensor transferability, suggesting that the utility of spectral normalisation is context- and sensor-dependent.

**Table 3.** Mean (standard deviation) F1-scores after 5-fold spatial cross-validation for LUCAS land cover classification. Results are shown for multispectral (Landsat-8), using Random Forest (RF). Ablation experiments are included where relevant. Bold values indicate the best-performing approach per block. For Experiment 3 (Wessels et al., 2016), performance was evaluated once on the full filtered validation set (rather than across the 5-fold LLTO splits), and therefore no standard deviation is reported.

| Exp. ID | Satellite Multispectral (Landsat-8) |
|---|---|
| | *Random Forest (RF)* |

| | | |
|---|---|---|
| **Gold standard** | 1 | **2015:** 0.69 (0.12) |
| | | **2018:** 0.69 (0.12) |
| **Naïve Baseline** | 2.1 | 0.69 (0.12) |
| **+ Normalisation** | 2.2 | 0.60 (0.10) |
| **Wessels et al. (2016)** | 3 | 0.68 |
| **Naïve Baseline** **+ Normalisation** **+ Manually labelled stable subset** | 4.1 | 0.70 (0.12) |
| **Naïve Baseline** **+ Normalisation** **+ CCD stable subset** | 4.2 | 0.69 (0.12) |
| **Manual relabelling (stage 1) + SSL (stage 2)** | 5.1 | **0.71 (0.11)** |
| **CCD (stage 1) + SSL (stage 2, Common Ground)** | 5.2 | **0.71 (0.10)** |

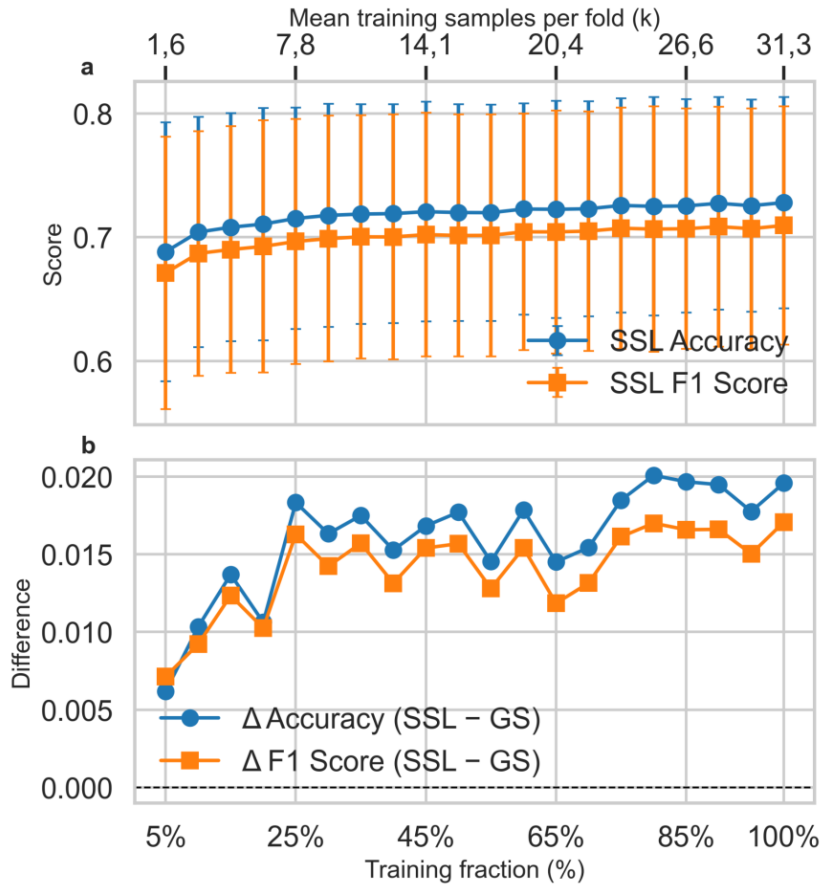

**Fig. 5.** *(a)* Semi-supervised learning (SSL) performance across increasing training sample sizes (5–100% or ~1600 - ~31300 samples) from case study 3 for LUCAS land cover. Accuracy and F1 scores are shown as mean ± standard deviation across 5 spatial cross-validation folds. *(b)* Difference in performance between SSL and the gold standard (fully supervised with complete relabelling) approach for the same training fractions. Positive values indicate that SSL outperforms the gold standard.

Taken together, the three case studies highlight the consistent advantages of combining unsupervised change detection (IRMAD/CCDC) with semi-supervised learning for Earth observation–based land cover and invasive species monitoring. While TabPFN demonstrated particular benefits in high-dimensional hyperspectral contexts, multispectral imagery performed similarly when paired with SSL, underscoring the practical utility of combining data from widely available sensors. L2 normalisation and cross-temporal training improved performance substantially in cross-sensor invasive species mapping, but these benefits did not generalise to the larger-scale LUCAS land cover application, where normalisation reduced accuracy. Notably, SSL yielded the greatest gains in data-sparse contexts with fine-class mapping targets, whereas in large-sample settings with broad class definitions the improvements were modest but still consistent. These results emphasise the importance of tailoring methodological choices to data dimensionality, sensor type, training set size and mapping targets.

## 4. Discussion

### 4.1 Principal Findings and Methodological Contributions

A central challenge in multi-temporal remote sensing is maintaining reliable classification accuracy across time while avoiding the substantial cost of collecting or relabelling reference data for each target time step (Gómez et al., 2016). This limitation constrains the scalability of ecosystem monitoring, particularly for applications such as land-cover change and invasive species mapping that require consistent labels across sensors, resolutions, and spatial extents. In this study, we demonstrate measurable progress toward addressing this challenge through a Common Ground framework that combines unsupervised change detection with semi-supervised learning (SSL) for systematic sample migration.

Across three case studies, our results provide consistent evidence that this approach yields a robust and scalable approach for updating reference datasets. Performance gains were observed across a wide range of conditions, including low- to high-spectral-resolution invasive species mapping and both local-extent and continental-scale multispectral land-cover classification. This consistency suggests that the framework generalizes beyond specific sensors, spatial scales, or thematic targets, extending the applicability of prior change-detection-based approaches.

The primary methodological contribution is a two-stage framework that directly addresses the temporal generalization problem in remote sensing classification. Stable areas identified via IRMAD or CCDC are treated as pseudo-ground truth, enabling reliable propagation of labels through time, while SSL is used to infer labels in spectrally dynamic regions. By decoupling stable and changing components of the landscape, the framework improves classification performance while removing the dependence on manual reannotation, thereby supporting more scalable and repeatable long-term monitoring workflows.

### 4.2. Performance Scaling and Data Efficiency

Our results indicate that the effectiveness of SSL depends less on sample size *per se* and more on the complexity of the mapping targets. In data-sparse regimes (~1,500 samples) with fine-class invasive tree species mapping targets, SSL provided notable improvements of up to 16% in F1-scores. Yet when the effect of sample size on accuracy was investigated with broad, land-cover targets, the difference in accuracy gains between data-sparse (~1500 samples) and data-rich settings (>33,000 samples) remained modest (0.5 - 2%), suggesting that SSL primarily supports learning of challenging target definitions rather than simply compensating for limited training data. This distinction has important operational implications: ecological monitoring programs often face constraints in both reference data availability and the difficulty of mapping fine-class targets, such as invasive species, and diverse land-cover types (Pereira et al., 2013; Pettorelli, 2017). By alleviating reliance on extensive ground surveys and enhancing performance where classification is inherently complex, SSL offers tangible benefits for organizations such as national park services, conservation NGOs, and developing country monitoring agencies (Olofsson et al., 2014; Tsendbazar et al., 2015).

### 4.3. Sensor Transferability and Operational Flexibility

The framework's robustness to sensor differences, with cross-sensor performance improvements of up to 30% when combined with L2 normalization and cross-temporal training, has important implications for long-term ecological monitoring. This capability becomes increasingly relevant as the remote sensing community transitions between satellite missions and integrates data from diverse sensors (Roy et al., 2014; Wulder, 2019). For example, we were able to improve the accuracy of invasive tree species mapping derived from NASA's BioSCape imaging spectroscopy mission with labels from a previous time step and different hyperspectral sensor. This demonstrates the method's ability to support continued mapping even when one sensor is discontinued and another begins operation, such as the transition from Landsat 8 to Landsat 9, or for experimental campaigns like BioSCape.

The differential response to normalization across case studies (beneficial for invasive species mapping using hyperspectral sensors, indifferent for invasive mapping with Sentinel-2, detrimental for LUCAS land cover) indicates that normalization strategies are beneficial for cross-sensor mapping but can be detrimental for same-sensor approaches. This finding highlights the importance of context-specific methodological choices and suggests that operational implementations should include systematic validation to determine when normalization improves versus degrades performance for specific sensor combinations and monitoring objectives.

### 4.4. Algorithmic Performance and High-Dimensional Data

TabPFN consistently outperformed Random Forest across our experiments, with the largest performance gains observed in hyperspectral contexts (up to 15 percentage point improvements in F1-scores). This pattern suggests that TabPFN's transformer-based architecture is particularly well-suited to high-dimensional spectral data, effectively modeling relationships that Random Forest, constrained by the curse of dimensionality, fails to capture (Belgiu & Drăguţ, 2016; Genuer et al., 2010; Hollmann et al., 2025). Despite these accuracy advantages, TabPFN's computational requirements present significant barriers to operational deployment. The model is constrained to a maximum of 10,000 training samples on consumer GPU hardware and requires approximately 11 days to process areas that Random Forest completes in 10 minutes—a ~1,600-fold increase in processing time. These limitations currently make TabPFN impractical for satellite-scale applications, though ongoing developments in transformer efficiency may eventually enable operational use (Hollmann et al., 2025).

### 4.5. Comparison to Existing Approaches

Our Common Ground framework's consistent performance advantages over the Wessels et al. (2016) approach (improvements of ~15%) can be attributed to three design characteristics that address limitations of previous methods. First, explicit change detection and pseudo-labeling reduces error propagation compared to direct map transfer. Second, inclusion of dynamic areas prevents bias against temporally variable land cover classes. Other sample migration approaches similar to Wessels likewise report declining performance at subsequent time steps (Ghorbanian et al., 2020; Huang et al., 2020), confirming a general limitation of methods that exclude dynamic areas when migrating samples. Third, cross-temporal training captures tempo-spectral distribution shifts more comprehensively than models restricted to single time periods. Notably, these improvements were achieved despite our framework operating with substantially (<2%) smaller training datasets (>3,000 samples over two time steps) compared to the Wessels et al. (2016) approach (162 294 samples), demonstrating that label quality and strategic temporal sampling are more important than label quantity alone. These improvements represent a methodological advance from temporal extrapolation to temporal invariance, where models adapt to temporal patterns rather than simply assuming historical relationships persist in future time steps (as is supported by the suboptimal performance of the naïve baseline, experiment 2).

### 4.6. Implications for Ecological Monitoring

Our results have implications for the resource requirements and temporal coverage of ecological monitoring programs where resources for repeated data collection are limited (Pereira et al., 2013; Scholes et al., 2012). By removing manual reannotation requirements at subsequent time steps while maintaining or improving classification accuracy, our approach lowers the cost and labour required for maintaining up-to-date land cover and invasive species maps, thereby enabling increased frequency and extent of monitoring activities. For instance, national biodiversity monitoring initiatives (Pettorelli, 2017), agencies running large-scale surveys such as the European LUCAS programme (Eurostat, 2020), or invasive species management authorities (van Wilgen et al., 2025) could use our approach to extend the value of existing datasets across multiple years without proportional increases in field survey effort, enabling more frequent land cover assessments for environmental policy and management applications (d'Andrimont et al., 2020; Eurostat, 2020). For invasive species monitoring in small areas such as protected areas, our approach makes it feasible for small research teams to monitor locally-relevant emerging invasions with limited or infrequent reference data recollection.

### 4.7. Limitations and Future Directions

Despite its strengths, the Common Ground approach is subject to several technical constraints. IRMAD is limited to bi-temporal change detection and is computationally intensive, which restricts its scalability (Canty & Nielsen, 2012). Although the pre-computed CCD product available on GEE offers greater scalability, it is only available for the period 2000–2019 (Z. Zhu & Woodcock, 2014). In this study, binary change/no-change masks derived from Landsat or Sentinel data were applied to the imaging spectroscopy data; this approach may introduce accuracy trade-offs that warrant further investigation. Additionally, the sample size limitation of TabPFN v2 (10,000 samples) and its relatively slow inference time currently preclude its use in near-real-time or time-sensitive mapping applications, despite its demonstrated accuracy advantages (Hollmann et al., 2025).

More fundamentally, our bi-temporal focus represents only the first step toward comprehensive multi-temporal analysis. Ecological monitoring requires understanding of gradual changes, seasonal cycles, and long-term trends that extend beyond simple before-after comparisons (Kennedy et al., 2010; Zhu & Woodcock, 2012). To improve the framework and cater for additional use cases, several research directions could be investigated. Robust learning methods may improve performance by filtering noisy labels or uncertain pseudo-labelled predictions. Multi-temporal extensions could examine performance degradation patterns and potential catastrophic forgetting effects. Testing shorter time intervals may reveal whether they may reduce spectral distribution shifts and improve temporal transferability (e.g. annual), or whether seasonality may create intra-annual differences that reduce transferability. Finally, regression applications for continuous variables like canopy height warrant investigation.

### 4.8. Recommendations

Based on our findings, we recommend several strategic directions for the remote sensing community. Firstly, investment in extending and updating the CCDC dataset beyond 2019 would provide critical infrastructure for SSL-based monitoring approaches (Zhe Zhu & Woodcock, 2014). Secondly, continued development of efficient foundation models specifically designed for Earth observation applications could unlock the accuracy potential demonstrated by TabPFN v2 while meeting operational constraints (Vaswani et al., 2017).

While datasets like LUCAS are invaluable for researchers, the results on the eight classes (reclassified from 82 classes) suggest that it might be reasonable to increase the interval between successive surveys. Moreover, labelling stable areas provides a high return on annotation investment when successfully used to provide pseudo-labels for dynamic areas. This finding could inform the design of future reference data collection campaigns and crowdsourcing initiatives (Fritz et al., 2015).

## 5. Conclusion

Our work demonstrates that integrating unsupervised change detection with semi-supervised learning creates a robust framework for temporal remote sensing that significantly reduces annotation burden while improving classification accuracy. The consistent performance improvements across diverse contexts—from hyperspectral invasive species mapping to continental-scale land cover classification—provide strong evidence for the approach's generalizability and operational potential. By addressing core challenges of error propagation, dynamic area exclusion, and temporal distribution shifts, this framework enables more cost-effective ecological monitoring. The introduced method offers a lightweight, computationally efficient alternative to manual relabelling or naïve temporal extrapolation, making it suitable for small to large scale monitoring use cases - increasingly important in an era of accelerating environmental change. The framework offers a practical solution for maintaining up-to-date environmental maps without proportional increases in field survey effort. This approach contributes to broader discussions in remote sensing and machine learning communities about label-efficient learning in dynamic environments, with applications extending beyond ecological monitoring to any domain where reference conditions change over time.

## 6. Code availability

All code used for data preprocessing, model training, evaluation, and figure generation is publicly available at https://doi.org/10.5281/zenodo.18479323. The repository includes a blogpost summarising this work with a code example, scripts and notebooks necessary to reproduce the experiments and results presented in this study.

7. **Data availability**

The Sentinel-2 Top-of-Atmosphere and Landsat-8 surface reflectance imagery used in this study are available through GEE as standard image collections. Reference datasets include the LUCAS land cover dataset and invasive species records, both accessed via GEE FeatureCollections. NASA's BioSCape airborne hyperspectral data is publicly available from the LP DAAC repository. Hyperspectral imagery provided by the Council for Geoscience is not publicly accessible.

8. **Declaration of generative AI and AI-assisted technologies in the manuscript preparation process**

During the preparation of this work the authors used ChatGPT and Claude in order to revise and improve the clarity of the manuscript text, and Gemini in order to generate the diagram in Figure A.1. After using these tools/services, the authors reviewed and edited the content as needed and take full responsibility for the content of the published article.

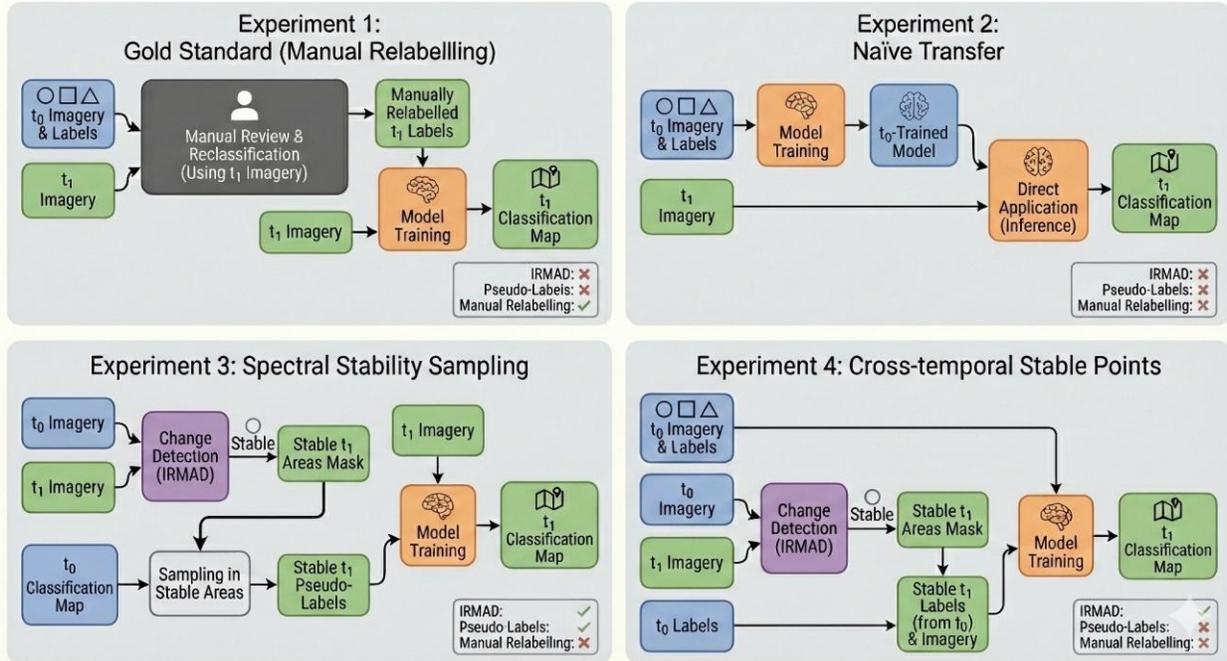

**Fig. A.1.** Overview of comparative experimental workflows to create a $t_1$ classification map. Blue and green components represent reference ($t_0$) and target ($t_1$) time steps, respectively. (1) Gold Standard: Supervised benchmark requiring manual reclassification of $t_0$ labels using $t_1$ imagery. (2) Naïve Transfer: Direct application of a model trained on $t_0$ data to $t_1$ imagery. (3) Spectral Stability Sampling: The $t_1$ model is trained on pseudo-labels sampled from the $t_0$ classification map within spectrally stable regions (identified via IRMAD or CCD). (4) Cross-temporal Stable Points: Training utilizes $t_0$ labels and images and original $t_0$ label coordinates located in spectrally stable regions, paired with corresponding $t_1$ spectral data. Refer to Figure 1 for the experiment 5 workflow, Common Ground.

**Table A.1.** Reference data allocation by experiment across all case studies. The table summarises how reference samples were defined, filtered, or augmented for each experimental configuration, including manual relabelling at $t_1$, stability-based filtering using binary change masks, and pseudo-label generation within a semi-supervised learning framework.

| Case study | Experiment | Reference data description |
|---|---|---|
| 1 & 2 | Exp. 1 | All $t_0$ (2018) reference points manually relabelled using $t_1$ (2023) imagery |
| 1 & 2 | Exp. 2 | All original $t_0$ (2018) reference points |
| 1 & 2 | Exp. 3 | Stable $t_0$–$t_1$ pixels identified; samples drawn from the $t_0$ (2018) classification map |
| 1 & 2 | Exp. 4.1 | Stable $t_0$–$t_1$ reference points filtered using manually derived binary change flags |
| 1 & 2 | Exp. 4.2 | Stable $t_0$–$t_1$ reference points filtered using IRMAD-derived binary change masks |
| 1 & 2 | Exp. 5.1 | Stable $t_0$–$t_1$ reference points filtered using manually derived binary change flags |
| 1 & 2 | Exp. 5.2 | Stable $t_0$–$t_1$ reference points filtered using IRMAD-derived binary change masks |
| 1 & 2 | Exp. 3 & 5 | Pseudo-labels generated for selected subsets of reference points |
| 3 | Exp. 1 | All $t_0$ (2015) reference points manually relabelled using $t_1$ (2018) imagery |
| 3 | Exp. 2 | All original $t_0$ (2015) reference points |
| 3 | Exp. 3 | Stable $t_0$–$t_1$ pixels identified using a CCDC-derived binary change mask; samples drawn from the $t_0$ (2015) classification map |
| 3 | Exp. 4.1 | Stable $t_0$–$t_1$ reference points filtered using manually derived binary change flags |
| 3 | Exp. 4.2 | Stable $t_0$–$t_1$ reference points filtered using CCDC-derived binary change masks applied to corresponding Landsat-8 imagery |
| 3 | Exp. 5.1 | Stable $t_0$–$t_1$ reference points filtered using manually derived binary change flags |
| 3 | Exp. 5.2 | Stable $t_0$–$t_1$ reference points filtered using CCDC-derived binary change masks |

| 3 | Exp. 3 & 5 | Pseudo-labels generated for selected subsets of reference points |

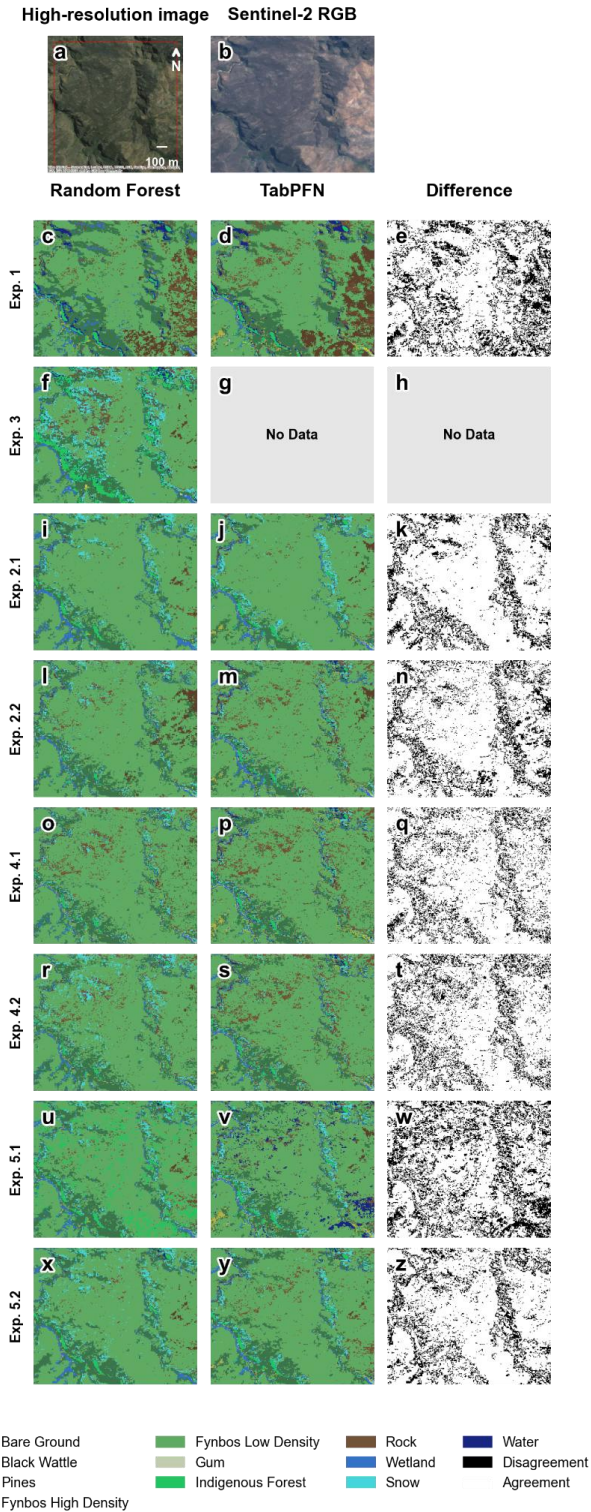

**Fig. A.2.** Qualitative comparison of predictions across the five experimental approaches for the multispectral invasive species case study. Panels show: (a) high-resolution ESRI imagery, (b) RGB Sentinel-2 imagery from 2023, and predictions from two models (Random Forest and TabPFN in columns 1 and 2, respectively) and their (dis)agreement (column 3). Specifically, (c-e) the gold-standard requiring complete relabelling, (f-h) Wessels et al. (2016), (i-k) naïve baseline, (l-n) naïve baseline with normalisation, (o-q) cross-temporal training with spectrally stable data identified using manually labelled change, (r-t) cross-temporal training with spectrally stable data identified using unsupervised change detection, (u-w) semi-supervised learning using manually labelled change, and (x-z) semi-supervised learning using unsupervised change detection. Note, there are no predictions for Exp. 3 (g-h), TabPFN since the number of pixels exceeded 10 000 and the subsequent prohibitively slow inference speeds.